# An evaluation of DeepSeek Models in Biomedical Natural Language Processing

Zaifu Zhan, MEng[1], Shuang Zhou, PhD[1], Huixue Zhou, MBBS[1], Jiawen Deng[1], Yu Hou, PhD[1], Jeremy Yeung, MS[1] and Rui Zhang, PhD[1]

[1]University of Minnesota Twin Cities, Minneapolis, MN, USA

**Abstract**
*The advancement of Large Language Models (LLMs) has significantly impacted biomedical Natural Language Processing (NLP), enhancing tasks such as named entity recognition, relation extraction, event extraction, and text classification. In this context, the DeepSeek series of models have shown promising potential in general NLP tasks, yet their capabilities in the biomedical domain remain underexplored. This study evaluates multiple DeepSeek models (Distilled-DeepSeek-R1 series and Deepseek-LLMs) across four key biomedical NLP tasks using 12 datasets, benchmarking them against state-of-the-art alternatives (Llama3-8B, Qwen2.5-7B, Mistral-7B, Phi-4-14B, Gemma-2-9B). Our results reveal that while DeepSeek models perform competitively in named entity recognition and text classification, challenges persist in event and relation extraction due to precision-recall trade-offs. We provide task-specific model recommendations and highlight future research directions. This evaluation underscores the strengths and limitations of DeepSeek models in biomedical NLP, guiding their future deployment and optimization.*

**Introduction**
Biomedical natural language processing (NLP)[1–3] plays a crucial role in modern healthcare, life sciences, and biomedical research. With the exponential growth of biomedical literature, electronic health records, radiology and pathology reports, clinical trial data, pharmaceutical and drug information, and medical dialogues, etc, NLP techniques have become indispensable for extracting valuable insights from unstructured text.[4,5] Effective biomedical NLP applications facilitate various tasks such as disease diagnosis,[6] drug discovery,[7] clinical decision support,[8] and knowledge graph construction.[9] These capabilities contribute to improved healthcare outcomes, accelerated research, and enhanced patient care.

However, biomedical NLP presents unique challenges that distinguish it from general-domain NLP.[10–12] First, biomedical texts contain highly specialized terminology, complex sentence structures, and numerous domain-specific abbreviations, making them difficult to process using conventional NLP models.[10,13] Unlike everyday language, where word meanings are generally well understood by pre-trained language models, biomedical terminology requires deep domain expertise, often necessitating external knowledge bases such as the Unified Medical Language System (UMLS).[12] Second, the availability of high-quality annotated datasets for training and fine-tuning biomedical NLP models is relatively limited compared to general-domain datasets.[14] While large-scale corpora exist for standard NLP tasks such as sentiment analysis or question answering, biomedical datasets often require manual curation by domain experts, making them costly and time-consuming to develop.[15] Additionally, biomedical language is highly dynamic, with continuous updates due to new medical discoveries, evolving clinical guidelines, and emerging diseases, posing challenges for model adaptation and knowledge retention.[16,17]

Over the past several years, large language models (LLMs)[18,19] have been successfully applied to various biomedical NLP tasks, leading to remarkable advancements in the field. For example, BiomedGPT,[20] a model fine-tuned on largescale medical corpora, has demonstrated strong capabilities in clinical text summarization, medical question-answering, and diagnostic assistance. Similarly, Gemini,[21] another cutting-edge model with multimodal capabilities, has shown promise in integrating text and medical imaging data for radiology report analysis, pathology classification, and automated medical documentation.[22] These successes highlight the transformative potential of LLMs in biomedical NLP, enabling applications ranging from automated literature review and clinical text mining to real-time diagnostic support and biomedical knowledge discovery.[14,23] These advancements have enabled various applications in the biomedical domain,[24,25] including assisting in clinical decision-making,[26] enhancing patient care,[27] and supporting medical research by providing accurate and contextually relevant information.[28–30] Their ability to process and interpret complex biomedical texts has also opened new possibilities for information extraction,[31] knowledge discovery,[23] and disease diagnosis,[6] significantly impacting the healthcare industry.

Recently, DeepSeek,[32–35] a series of foundation models trained on large-scale multilingual and multimodal data, has gained significant attention for its strong performance across diverse NLP tasks. DeepSeek models leverage advanced architectures, extensive training corpora, and innovative reinforcement learning techniques, making them competitive with, or even surpassing, state-of-the-art models such as GPT-4 and Llama 3 in various benchmarks. Among Deepseek family, models like DeepSeekLLM[34] and the distilled DeepSeek-R1[32] are of significant research value. These models are not excessively large, making them accessible for fine-tuning by individual researchers for applications in biomedical domains.[36] The distilled versions, in particular, offer a balance between performance and resource efficiency, enabling their deployment on standard computational setups commonly available in research institutions.[37] Moreover, the open-source nature of these models fosters collaboration and innovation within the scientific community, as researchers can share improvements and insights, collectively enhancing the models' capabilities and broadening their applicability across various biomedical challenges.

Despite their impressive general-domain performance, DeepSeek models' effectiveness in biomedical NLP remains largely unexplored. To fill this gap, we conducted a thorough evaluation of the distilled DeepSeek-R1 and DeepSeek-LLM models within the biomedical domain. Specifically, we assessed their performance across 4 common tasks with 12 datasets, benchmarked it against the current state-of-the-art models (Mistral,[38] Llama3,[39] Qwen2.5,[40] Phi-4,[41] and Gemma-2[21]), and analyzed its strengths and limitations. Through this evaluation, we seek to provide a comprehensive understanding of DeepSeek's capabilities and offer insights into its potential applications in biomedical NLP.

## Methods

*Tasks and datasets*

In this study, we conducted a comprehensive evaluation of DeepSeek models across four pivotal natural language processing tasks: event extraction, relation extraction, named entity recognition, and text classification. Event extraction identifies specific occurrences in text, such as biological or medical events, using datasets like PHEE[42] (pharmacovigilance events), Genia2011[43] (10 biomedical event types), and Genia2013[44] (12 refined event types). Relation extraction focuses on entity interactions, employing datasets like DDI[45] (drug-drug interactions), GIT[46] (22 general biomedical relations), and BioRED[47] (8 biological relation types). Named entity recognition classifies entities into predefined categories using BC5CDR[48] (chemicals and diseases), BC2GM[49] (gene mentions), and BC4Chemd[50] (chemical and drug names). Text Classification assigns predefined labels to text segments, utilizing ADE[51] (adverse drug events), PubMed20k RCT[52] (biomedical literature categorization), and HealthAdvice[53] (health-related advice classification). These datasets support a comprehensive evaluation of DeepSeek models across multiple NLP tasks. The statistics of selected datasets are shown in Table. 1

**Table 1.** Dataset statistics for four tasks

| Task | Dataset | Train | Validation | Test |
|---|---|---|---|---|
| Event Extraction | PHEE[42] | 2,898 | 961 | 968 |
| | Genia2011[43] | 8,666 | 249 | 2,886 |
| | Genia2013[44] | 222 | 249 | 305 |
| Relation Extraction | DDI[45] | 11,556 | 1,285 | 3,020 |
| | GIT[46] | 3,731 | - | 465 |
| | BioRED[47] | 400 | 100 | 100 |
| Named Entity Recognition | BC5CDR[48] | 5,226 | 5,330 | 5,865 |
| | BC2GM[49] | 12,500 | 2,500 | 5,000 |
| | BC4ChemD[50] | 30,683 | 3,064 | 26,365 |
| Text Classification | ADE[51] | 18,812 | - | 4,704 |
| | PubMed20k RCT[52] | 180,040 | 30,212 | 30,135 |
| | HealthAdvice[53] | 6,940 | - | 1,736 |

*Selected DeepSeek models and Baselines*

Our primary focus is on the distilled DeepSeek-R1 series of models,[32] which have been distilled to enhance efficiency while maintaining high performance. These models include DeepSeek-R1-Distill-Llama-70B, DeepSeek-R1-Distill-Qwen-32B, DeepSeek-R1-Distill-Qwen-14B, DeepSeek-R1-Distill-Llama-8B, and DeepSeek-R1-Distill-Qwen-7B. The distillation process employed in these models effectively compresses the knowledge and capabilities of larger models into more compact forms, resulting in improved computational efficiency without significant loss in performance. This makes them particularly suitable for applications where resource constraints are a consideration.

Additionally, we have included the Deepseek-LLM-67B-base and Deepseek-LLM-7B-base models[34] in our evaluation. These models, comprising 67 billion and 7 billion parameters respectively, have been trained from scratch on a vast dataset of 2 trillion tokens in both English and Chinese. The Deepseek-LLM-67B-base model has demonstrated superior general capabilities, outperforming models like Llama2-70B Base in areas such as reasoning, coding, mathematics, and Chinese comprehension. The Deepseek-LLM-7B-base model, while smaller in scale, maintains robust performance across various tasks, making it a viable option for scenarios where computational resources are limited.

As baselines, we have selected the Llama3-8B,[39] Qwen2.5-7B,[40] Mistral-7B-Instruct-v0.2,[38] Phi-4,[41] Gemma-2-9B,[21] models. These models are not only foundational architectures from which some of the DeepSeek-R1 models have been distilled but are also widely recognized open-source models in the research community. Their open-source nature is especially pertinent in biomedical research settings, such as hospitals and research institutions, where data privacy and security are paramount. The availability of open-source models allows for greater flexibility and control, facilitating their deployment in environments with stringent privacy requirements.

By comparing the performance of the distilled DeepSeek-R1 models and the DeepSeek-LLM base models against these established baselines, we aim to gain comprehensive insights into their relative strengths and suitability for various tasks. This evaluation will inform the selection of appropriate models for specific applications, particularly in contexts where computational resources and data privacy are critical considerations.

*Metrics*

In this study, we employ precision, recall, and F1-score as evaluation metrics across all tasks. These metrics are calculated using the exact match approach, meaning that a prediction is considered correct only if it exactly matches the ground truth. By utilizing the exact match criterion, we ensure that only predictions that fully align with the ground truth are counted as correct, thereby maintaining stringent evaluation standards.

*Experimental setting*

We utilized NVIDIA A100 GPUs equipped with 40GB of memory, providing ample computational resources for both training and inference. We set the batch size to 4 per device during both training and evaluation phases. Inference was conducted on a per-sentence basis to enhance efficiency. For automated evaluation, we employed Low-Rank Adaptation (LoRA),[54] a parameter-efficient fine-tuning technique. Specifically, we configured LoRA with a rank of 64, an alpha value of 32, and a dropout rate of 0.1 to fine-tune the model. The model optimization was performed using the AdamW optimizer with a learning rate of 1e-5. The fine-tuning process spanned 5,000 steps, with evaluations occurring every 1,000 steps. The model demonstrating the best performance during these evaluations was subsequently selected for inference.

**Results**

We present all the evaluation results in Fig. 1 and readers can find all the numeric values for each task in the appendix. Overall, the results demonstrate that while modern LLMs have reached strong levels of proficiency in named entity recognition and text classification, challenges persist in event and relation extraction. The trade-offs between precision and recall are particularly evident in extraction tasks, indicating that further fine-tuning and model adaptation are necessary for improved accuracy in biomedical applications.

*Event extraction:* Mistral-7B-Instruct-v0.2, DeepSeek-R1-Distill-Llama-70B, and DeepSeek-R1-Distill-Qwen-14B achieved the highest F1 scores on the PHEE dataset, all exceeding 0.95. Qwen2.5-7B-Instruct, Phi-4, and Gemma-2-9B also performed well, with F1 scores above 0.94. However, performance on the Genia2013 dataset was significantly lower across all models, with the highest F1 score reaching only 0.3348 by Deepseek-LLM-7B-base. Llama3-8B demonstrated high recall on PHEE and Genia2011 but had comparatively lower precision, suggesting that while it identifies a large number of event mentions, it does so at the cost of accuracy.

*Relation extraction:* Models exhibited more variability in performance across the DDI, GIT, and BioRED datasets. Mistral-7B-Instruct-v0.2, DeepSeek-R1-Distill-Llama-70B, and DeepSeek-R1-Distill-Qwen-32B had the strongest F1 scores across all three datasets, exceeding 0.76. Llama3-8B showed the highest recall in BioRED (0.9595) but with much lower precision, leading to an F1 score of 0.5352. DeepSeek-R1-Distill-Qwen-14B and Phi-4 also demonstrated balanced performance, with F1 scores in the range of 0.74–0.77 across the datasets. The observed variability indicates that while some models are optimized for high recall, they may introduce more false positives, whereas others achieve a better balance between precision and recall.

*Named entity recognition:* The models performed consistently well across the BC5CDR, BC2GM, and BC4Chemd datasets. Llama3-8B achieved the highest F1 score on BC4Chemd (0.9791) and remained among the top-performing models for the other two datasets. DeepSeek-R1-Distill-Llama-70B also demonstrated strong and stable performance, with F1 scores above 0.96 across datasets. Most models, including Phi-4, Gemma-2-9B, and DeepSeek-R1-Distill-Qwen-32B, achieved F1 scores above 0.95, indicating that NER is a well-handled task among these models. The slight variations in recall suggest that while most models reliably detect common biomedical entities, some may struggle with rare or complex entity mentions.

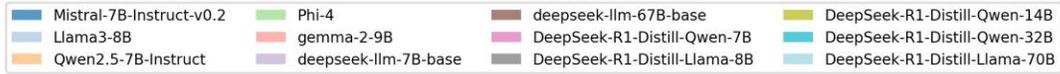
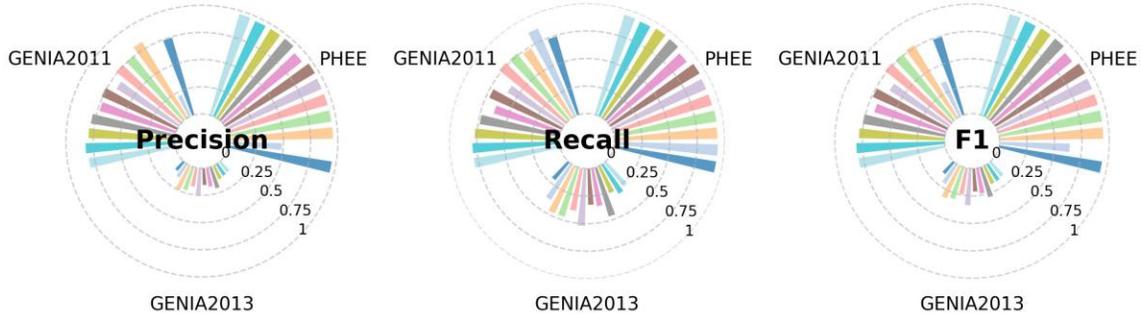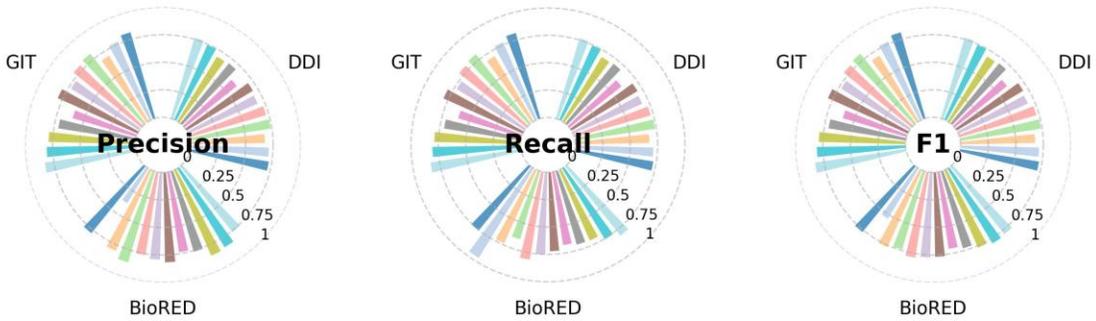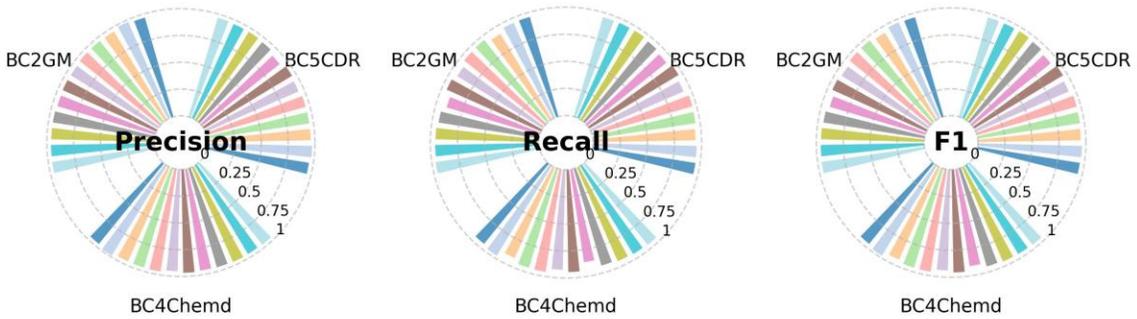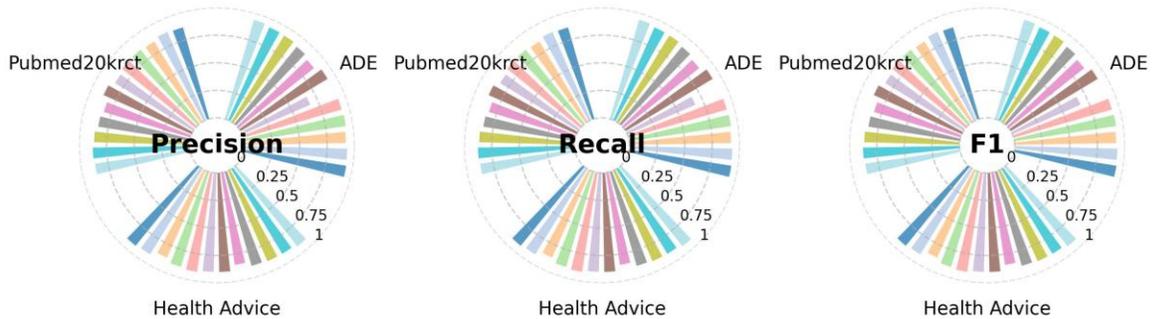

**Figure 1.** Evaluation results for 12 LLMs on 4 tasks across 12 datasets using precision, recall, F1 as the metrics

*Text classification:* Performance remained relatively high across the ADE, PubMed 20k RCT, and HealthAdvice datasets. Mistral-7B-Instruct-v0.2 and DeepSeek-R1-Distill-Llama-70B obtained the highest F1 scores, consistently reaching or exceeding 0.937. Llama3-8B, Phi-4, and Gemma-2-9B also performed well, with F1 scores ranging from 0.91 to 0.98. However, Deepseek-LLM-7B-base showed significantly lower performance on the ADE dataset, with an F1 score of 0.673. This suggests that while most models generalize well across classification tasks, certain datasets may pose challenges depending on the model's training corpus and optimization strategy.

**Discussion**

Evaluating state-of-the-art large language models (LLMs) in biomedical NLP is crucial for understanding their capabilities and limitations. With rapid advancements in transformer-based architecture and instruction-tuned models, it is essential to benchmark their performance across diverse datasets to assess generalizability, robustness, and efficiency. Such evaluations provide insights into whether these models can replace traditional rule-based or supervised learning methods, guide the selection of models for real-world applications, and identify areas that require further adjustment. Additionally, testing on biomedical datasets helps uncover whether these models have effectively learned specialized medical knowledge from their training corpora, which is critical for ensuring their reliability in fields like clinical decision support, drug discovery, and scientific literature analysis.

Our evaluation across event extraction, relation extraction, named entity recognition, and text classification reveals significant differences in model performance based on task complexity and dataset characteristics. While some models, such as DeepSeekR1-Distill-Llama-70B and Mistral-7B-Instruct-v0.2, consistently perform well across tasks, others exhibit varying strengths and weaknesses depending on the task's requirements. The results highlight the progress LLMs have made in biomedical NLP, but they also underscore persistent challenges, particularly in tasks requiring fine-grained semantic understanding, such as event and relation extraction.

A key finding in event extraction is the substantial variation in performance between datasets. Most models achieve high accuracy on PHEE but struggle significantly on Genia2013, indicating that dataset-specific factors, such as annotation style, complexity of event structures, or entity ambiguity, can impact extraction performance. This suggests that while LLMs have learned to recognize simple event structures, they still struggle with more nuanced or multi-step events, particularly those requiring deeper contextual reasoning. The observed drop in precision for models with high recall, such as Llama3-8B, suggests that some models lean towards aggressive event detection at the cost of accuracy. This trade-off is crucial in biomedical applications, where false positives can lead to misleading conclusions.

For relation extraction, models generally perform better than in event extraction but still exhibit significant differences in precision and recall trade-offs. The observation that Llama3-8B has an exceptionally high recall but lower precision on BioRED suggests that certain models prioritize exhaustive relation discovery, which may be beneficial in exploratory settings but problematic in high-precision applications such as clinical decision support. Conversely, models like DeepSeek-R1-DistillQwen-32B and DeepSeek-R1-Distill-Qwen-14B maintain more balanced precision and recall, making them better suited for structured knowledge extraction tasks. The results imply that relation extraction remains a challenging task for LLMs, particularly in complex biomedical literature, where relationships are often implied rather than explicitly stated.

In contrast, NER shows relatively stable performance across datasets and models, suggesting that biomedical entity recognition is a task that LLMs handle well, likely due to the availability of large-scale labeled training data. The high F1 scores achieved by multiple models indicate that most of them have successfully learned the necessary linguistic patterns to identify biomedical entities with high accuracy. However, the slight drop in recall for certain models suggests that some may be more conservative in entity recognition, potentially missing rare or novel terms. Given that entity recognition serves as the foundation for more complex NLP tasks like event and relation extraction, continued improvements in handling ambiguous or multi-word entity mentions could further enhance downstream task performance.

Text classification emerges as the most stable task across models, with relatively small performance variations. Unlike extraction-based tasks, where the complexity of linguistic structures plays a major role, text classification likely benefits from clearer input-output mappings, making it easier for models to generalize. The consistently high F1 scores across datasets suggest that current LLMs are well-equipped for text classification tasks in biomedical domains. However, Deepseek-LLM-7B-base's lower performance on the ADE dataset indicates that some models may still struggle with specific classification tasks, possibly due to differences in training corpus exposure. This suggests that while LLMs can be readily deployed for classification tasks, dataset-specific fine-tuning may still be necessary for optimal results.

*Model Recommendations*

Based on performance variations across tasks, we provide targeted model recommendations.

- For event extraction, Mistral-7B-Instruct-v0.2 and DeepSeek-R1-Distill-Llama-70B offer the best balance of precision and recall, making them reliable for structured event extraction in biomedical texts. While Llama3-8B achieves high recall, its tendency for false positives limits its use in precision-critical tasks.

- For relation extraction, DeepSeek-R1-Distill-Qwen-32B and DeepSeek-R1-Distill-Qwen-14B excel in balanced extraction, while Llama3-8B is preferable for high-recall tasks like hypothesis generation. Mistral-7B-Instruct-v0.2 and DeepSeek-R1Distill-Llama-70B also perform well but may not be ideal when exceptionally high recall is needed.

- For named entity recognition (NER), most models perform well. Llama3-8B achieves the highest accuracy, while DeepSeekR1-Distill-Llama-70B and Mistral-7B-Instruct-v0.2 offer consistently strong performance. Given NER's robustness across models, selection should consider efficiency and deployment needs over accuracy alone.

- For text classification, Mistral-7B-Instruct-v0.2, DeepSeek-R1-Distill-Llama-70B, and Llama3-8B provide stable, high F1 scores. However, Deepseek-LLM-7B-base underperforms on the ADE dataset, making it less suitable for high-accuracy classification tasks.

Overall, DeepSeek-R1-Distill-Llama-70B and Mistral-7B-Instruct-v0.2 are strong generalists, while Llama3-8B suits recall-intensive tasks and DeepSeek-R1-Distill-Qwen-32B is ideal for structured relation extraction. Model choice should balance precision, recall, and computational efficiency based on task needs.

*Future Directions*
Despite advancements in biomedical NLP, challenges remain in event and relation extraction. Future research could explore retrieval-augmented generation (RAG)[55–57] and chain-of-thought (CoT)[58] reasoning to enhance accuracy, robustness, and interpretability.
RAG enables models to retrieve relevant biomedical knowledge dynamically, addressing issues of outdated or incomplete information. It could improve event extraction by fetching supporting evidence and relation extraction by grounding predictions in factual data. Efficient real-time retrieval integration remains a key research area.
CoT reasoning helps models break down complex tasks into intermediate steps, improving structured information extraction. It could enhance causality understanding in event and relation extraction, especially when combined with self-consistency techniques for more reliable outputs.
Integrating RAG for dynamic knowledge retrieval and CoT for structured reasoning could significantly improve biomedical NLP models for applications in medical research, clinical decision support, and automated literature analysis.

**Conclusion**
This study presents a comprehensive evaluation of DeepSeek models in biomedical NLP, benchmarking their performance against established state-of-the-art alternatives. Our findings indicate that while these models excel in named entity recognition and text classification, challenges persist in event and relation extraction, particularly in maintaining a balance between precision and recall. Among the evaluated models, DeepSeek-R1-Distill-Llama-70B and Mistral-7B-Instruct-v0.2 emerge as strong generalists, whereas DeepSeek-R1-Distill-Qwen-32B demonstrates particular effectiveness in structured relation extraction. These insights inform model selection for real-world biomedical applications, particularly in domains requiring high accuracy and computational efficiency.

**Acknowledgements**
This work was supported by the National Institutes of Health's National Center for Complementary and Integrative Health under grant numbers R01AT009457 and U01AT012871, the National Institute on Aging under grant number R01AG078154, the National Cancer Institute under grant number R01CA287413, the National Institute of Diabetes and Digestive and Kidney Diseases under grant number R01DK115629, and the National Institute on Minority Health and Health Disparities under grant number 1R21MD019134-01.

**Appendix**

**Table 2.** Evaluation results for event extraction datasets. P and R represent precision and recall scores, respectively.

| Model | PHEE | | | Genia2011 | | | Genia2013 | | |
|---|---|---|---|---|---|---|---|---|---|
| | P | R | F1 | P | R | F1 | P | R | F1 |
| Mistral-7B-Instruct-v0.2 | 0.9587 | 0.9355 | 0.9469 | 0.741 | 0.7391 | 0.74 | 0.1036 | 0.2066 | 0.138 |
| Llama3-8B | 0.4851 | 0.9375 | 0.6394 | 0.2156 | 0.8656 | 0.3452 | 0.1442 | 0.3754 | 0.2083 |
| Qwen2.5-7B-Instruct | 0.9556 | 0.9325 | 0.9439 | 0.8159 | 0.7324 | 0.7719 | 0.2444 | 0.4689 | 0.3213 |
| Phi-4 | 0.9538 | 0.9375 | 0.9456 | 0.7543 | 0.7676 | 0.7609 | 0.2233 | 0.4656 | 0.3018 |
| Gemma-2-9B | 0.9535 | 0.9304 | 0.9418 | 0.7624 | 0.7727 | 0.7675 | 0.1747 | 0.3902 | 0.2414 |
| Deepseek-LLM-7B-base | 0.9525 | 0.9294 | 0.9408 | 0.6708 | 0.5957 | 0.6311 | 0.2471 | 0.519 | 0.3348 |
| Deepseek-LLM-67B-base | 0.9629 | 0.9415 | 0.9521 | 0.7591 | 0.7198 | 0.7389 | 0.1548 | 0.3279 | 0.2103 |
| DeepSeek-R1-Distill-Qwen-7B | 0.9249 | 0.9062 | 0.9155 | 0.7311 | 0.6317 | 0.6778 | 0.1755 | 0.3475 | 0.2332 |
| DeepSeek-R1-Distill-Llama-8B | 0.9638 | 0.9405 | 0.952 | 0.7828 | 0.7349 | 0.7581 | 0.2032 | 0.4623 | 0.2823 |
| DeepSeek-R1-Distill-Qwen-14B | 0.9611 | 0.9476 | 0.9543 | 0.7931 | 0.7685 | 0.7806 | 0.1348 | 0.2656 | 0.1788 |
| DeepSeek-R1-Distill-Qwen-32B | 0.9588 | 0.9375 | 0.948 | 0.8195 | 0.7886 | 0.8038 | 0.1238 | 0.3082 | 0.1767 |
| DeepSeek-R1-Distill-Llama-70B | 0.969 | 0.9456 | 0.9571 | 0.8019 | 0.7878 | 0.7948 | 0.1183 | 0.2754 | 0.1655 |

**Table 3.** Evaluation results for relation extraction datasets. P and R represent precision and recall scores, respectively.

| Model | DDI | | | GIT | | | BioRED | | |
|---|---|---|---|---|---|---|---|---|---|
| | P | R | F1 | P | R | F1 | P | R | F1 |
| Mistral-7B-Instruct-v0.2 | 0.722 | 0.722 | 0.722 | 0.8129 | 0.8129 | 0.8129 | 0.7814 | 0.7568 | 0.7689 |
| Llama3-8B | 0.711 | 0.711 | 0.711 | 0.772 | 0.772 | 0.772 | 0.3711 | 0.9595 | 0.5352 |
| Qwen2.5-7B-Instruct | 0.675 | 0.675 | 0.675 | 0.6946 | 0.6946 | 0.6946 | 0.803 | 0.7342 | 0.7671 |
| Phi-4 | 0.738 | 0.738 | 0.738 | 0.8065 | 0.8065 | 0.8065 | 0.8623 | 0.6486 | 0.7404 |
| Gemma-2-9B | 0.724 | 0.724 | 0.724 | 0.7914 | 0.7914 | 0.7914 | 0.7627 | 0.8108 | 0.786 |
| Deepseek-LLM-7B-base | 0.687 | 0.687 | 0.687 | 0.7269 | 0.7269 | 0.7269 | 0.7915 | 0.7523 | 0.7714 |
| Deepseek-LLM-67B-base | 0.703 | 0.703 | 0.703 | 0.7935 | 0.7935 | 0.7935 | 0.8163 | 0.7207 | 0.7656 |
| DeepSeek-R1-Distill-Qwen-7B | 0.607 | 0.607 | 0.607 | 0.6043 | 0.6043 | 0.6043 | 0.7353 | 0.6757 | 0.7042 |
| DeepSeek-R1-Distill-Llama-8B | 0.702 | 0.702 | 0.702 | 0.7097 | 0.7097 | 0.7097 | 0.7598 | 0.6982 | 0.7277 |
| DeepSeek-R1-Distill-Qwen-14B | 0.69 | 0.69 | 0.69 | 0.7871 | 0.7871 | 0.7871 | 0.8478 | 0.7027 | 0.7685 |
| DeepSeek-R1-Distill-Qwen-32B | 0.744 | 0.744 | 0.744 | 0.8043 | 0.8043 | 0.8043 | 0.8358 | 0.7568 | 0.7943 |
| DeepSeek-R1-Distill-Llama-70B | 0.764 | 0.764 | 0.764 | 0.8323 | 0.8323 | 0.8323 | 0.7627 | 0.8108 | 0.786 |

**Table 4.** Evaluation results for named entity recognition datasets. P and R represent precision and recall scores, respectively.

| Model | BC5CDR | | | BC2GM | | | BC4Chemd | | |
|---|---|---|---|---|---|---|---|---|---|
| | P | R | F1 | P | R | F1 | P | R | F1 |
| Mistral-7B-Instruct-v0.2 | 0.9588 | 0.9581 | 0.9584 | 0.9645 | 0.9569 | 0.9606 | 0.9663 | 0.9511 | 0.9586 |
| Llama3-8B | 0.974 | 0.9546 | 0.9642 | 0.9801 | 0.956 | 0.9678 | 0.986 | 0.972 | 0.9791 |
| Qwen2.5-7B-Instruct | 0.9601 | 0.9444 | 0.9522 | 0.96 | 0.9219 | 0.9405 | 0.9624 | 0.9143 | 0.9377 |
| Phi-4 | 0.9597 | 0.9574 | 0.9585 | 0.9628 | 0.96 | 0.9614 | 0.9656 | 0.9536 | 0.9596 |
| Gemma-2-9B | 0.961 | 0.9633 | 0.9622 | 0.9629 | 0.9569 | 0.9599 | 0.9681 | 0.9564 | 0.9623 |
| Deepseek-LLM-7B-base | 0.9471 | 0.9323 | 0.9396 | 0.9543 | 0.9245 | 0.9391 | 0.9503 | 0.9236 | 0.9368 |
| Deepseek-LLM-67B-base | 0.96 | 0.9614 | 0.9607 | 0.9567 | 0.9534 | 0.9551 | 0.9652 | 0.9446 | 0.9548 |
| DeepSeek-R1-Distill-Qwen-7B | 0.9267 | 0.9329 | 0.9298 | 0.9506 | 0.8969 | 0.923 | 0.9574 | 0.8662 | 0.9095 |
| DeepSeek-R1-Distill-Llama-8B | 0.9557 | 0.9504 | 0.953 | 0.9584 | 0.9384 | 0.9483 | 0.9614 | 0.9306 | 0.9457 |
| DeepSeek-R1-Distill-Qwen-14B | 0.9624 | 0.9617 | 0.962 | 0.9592 | 0.9529 | 0.9561 | 0.9655 | 0.9435 | 0.9544 |
| DeepSeek-R1-Distill-Qwen-32B | 0.9616 | 0.9593 | 0.9604 | 0.9616 | 0.9561 | 0.9588 | 0.9635 | 0.9505 | 0.957 |
| DeepSeek-R1-Distill-Llama-70B | 0.9657 | 0.9571 | 0.9614 | 0.9635 | 0.9645 | 0.964 | 0.9684 | 0.957 | 0.9627 |

**Table 5.** Evaluation results for text classification datasets. P and R represent precision and recall scores, respectively.

| Model | ADE | | | Pubmed20k RCT | | | Health Advice | | |
|---|---|---|---|---|---|---|---|---|---|
| | P | R | F1 | P | R | F1 | P | R | F1 |
| Mistral-7B-Instruct-v0.2 | 0.937 | 0.937 | 0.937 | 0.864 | 0.864 | 0.864 | 0.922 | 0.922 | 0.922 |
| Llama3-8B | 0.932 | 0.932 | 0.932 | 0.873 | 0.873 | 0.873 | 0.91 | 0.91 | 0.91 |
| Qwen2.5-7B-Instruct | 0.92 | 0.92 | 0.92 | 0.85 | 0.85 | 0.85 | 0.864 | 0.864 | 0.864 |
| Phi-4 | 0.932 | 0.932 | 0.932 | 0.859 | 0.859 | 0.859 | 0.9 | 0.9 | 0.9 |
| Gemma-2-9B | 0.927 | 0.927 | 0.927 | 0.855 | 0.855 | 0.855 | 0.912 | 0.912 | 0.912 |
| Deepseek-LLM-7B-base | 0.673 | 0.673 | 0.673 | 0.833 | 0.833 | 0.833 | 0.898 | 0.898 | 0.898 |
| Deepseek-LLM-67B-base | 0.9213 | 0.9213 | 0.9213 | 0.874 | 0.874 | 0.874 | 0.8992 | 0.8992 | 0.8992 |
| DeepSeek-R1-Distill-Qwen-7B | 0.876 | 0.876 | 0.876 | 0.819 | 0.819 | 0.819 | 0.846 | 0.846 | 0.846 |
| DeepSeek-R1-Distill-Llama-8B | 0.901 | 0.901 | 0.901 | 0.839 | 0.839 | 0.839 | 0.888 | 0.888 | 0.888 |
| DeepSeek-R1-Distill-Qwen-14B | 0.919 | 0.919 | 0.919 | 0.864 | 0.864 | 0.864 | 0.897 | 0.897 | 0.897 |
| DeepSeek-R1-Distill-Qwen-32B | 0.922 | 0.922 | 0.922 | 0.875 | 0.875 | 0.875 | 0.899 | 0.899 | 0.899 |
| DeepSeek-R1-Distill-Llama-70B | 0.938 | 0.938 | 0.938 | 0.867 | 0.867 | 0.867 | 0.921 | 0.921 | 0.921 |